\documentclass{article}
\pdfoutput=1

\PassOptionsToPackage{numbers}{natbib}

\usepackage[final]{nips_2017_mod}

\usepackage[utf8]{inputenc} 
\usepackage[T1]{fontenc}    
\usepackage{hyperref}       
\usepackage{url}            
\usepackage{booktabs}       
\usepackage{amsfonts}       
\usepackage{nicefrac}       
\usepackage{microtype}      

\usepackage{amsmath}
\usepackage{amssymb}
\usepackage{multirow}
\usepackage{float}
\usepackage{tikz}
\usepackage{graphicx}
\usepackage{caption}
\usepackage{subcaption}
\usetikzlibrary{shapes,arrows}
\usetikzlibrary{calc}

\title{Second-Order Word Embeddings\\from Nearest Neighbor Topological Features}

\author{
  Denis R.~Newman-Griffis$^{\spadesuit,\clubsuit}$ and Eric~Fosler-Lussier$^{\spadesuit}$ \\
  $^\spadesuit$The Ohio State University, Columbus, OH, USA\\
  $^\clubsuit$National Institutes of Health, Clinical Center, Bethesda, MD, USA\\
  {\tt newman-griffis.1@osu.edu, fosler@cse.ohio-state.edu}
}

\begin{document}

\maketitle

\begin{abstract}
We introduce second-order vector representations of words,
induced from nearest neighborhood topological features in pre-trained contextual
word embeddings.
We then analyze the effects of using second-order embeddings as input features
in two deep natural language processing models, for named entity recognition
and recognizing textual entailment, as well as a linear model for paraphrase
recognition.  Surprisingly, we find that nearest neighbor information alone
is sufficient to capture most of the performance benefits derived from using
pre-trained word embeddings.  Furthermore, second-order embeddings are able to handle
highly heterogeneous data better than first-order representations, though at the cost of some
specificity.  Additionally, augmenting contextual embeddings with second-order information
further improves model performance in some cases.
Due to variance in the random initializations of word embeddings, utilizing nearest neighbor
features from multiple first-order embedding samples can also contribute to downstream
performance gains.  Finally, we identify intriguing characteristics of second-order
embedding spaces for further research, including much higher density and
different semantic interpretations of cosine similarity.
\end{abstract}

\section{Introduction}
\label{sec:introduction}

Word embeddings are dense, low-dimensional vector representations of words that are commonly used as input features in a
variety of natural language processing (NLP) tasks \cite{Turney2010}.  In contrast to symbolic one-hot or hierarchical clustering--based
representations, real-valued embedding vectors easily reflect varying degrees of similarity between words, and significantly
reduce sparsity in linear algebra operations.  The most common methods for learning word embeddings take an unsupervised 
approach based on word cooccurrence, using fixed-width context windows within a large text corpus \cite{Bengio2003,
Collobert2011,Mikolov2013a,Pennington2014}.  Compressing this contextual information via neural language modeling gives
representations that retain some semantic and syntactic properties of the target words \cite{Mikolov2013c,Hamilton2016,Avraham2017},
and often lead to large performance gains when used as input features in downstream NLP systems \cite{Turian2010,
Melamud2016}.

There has been significant recent research on improving the utility of word embeddings as downstream features, by modifying
the contextual information used in training \cite{Levy2014c,Schwartz2016}, by augmenting the embeddings with additional data
relevant to the target task \cite{Faruqui2015,KimJK2016}, or both \cite{Vulic2017}.  A number of other recent studies have
approached embeddings
from another angle, by trying to analyze exactly what is encoded in the space characterized by the embedded representations
\cite{Fyshe2015,Li2016,Li2017a}.
Since embeddings are usually initialized to random locations in the target $d$-dimensional space, and then trained based on
position relative to observed context words, values of individual features are notoriously difficult to interpret.  However,
a recent study by Linzen \cite{Linzen2016} illustrated that in some semantic tasks, what matters most is neighborhood
structure in the embedding space.  Thus, while absolute position is clearly informative, we hypothesize that the
{\it relative} position of words encodes the most critical information for downstream tasks.

In this work, we present a novel method for deriving second-order word representations from the nearest neighborhood topology of
pre-trained word embeddings, and analyze the results of using these representations in downstream NLP applications.
In particular, we propose a two-step process of using inducing a
$k$--nearest neighbor graph from pre-trained embeddings,\footnote{
    We use cosine similarity to calculate nearest neighbors; all further references to distance or similarity in this
    paper refer to cosine distance or similarity.
} where each node is a word, and then using recent methods for
unsupervised graph embeddings to re-learn word representations from this graph.  We explore using these second-order embeddings
in proven models for three downstream tasks: named entity recognition, textual entailment, and
paraphrase recognition.  We find that replacing first-order contextual embeddings with second-order embeddings as input features
yields almost equivalent performance to the original word embeddings, and actually increases recall in some cases.
Furthermore, we show that augmenting first-order embeddings with second-order information can improve performance when used in
non-linear models, especially on heterogeneous data, although the additional information confuses linear models.  Finally, we
analyze the changes in the nearest neighborhoods of selected terms between first-order and second-order embeddings, and find
that the second-order space is significantly denser than the original contextual embedding space.
\section{Related Work}
\label{sec:related_work}

\subsection{Alternate features for training embeddings}

Recent research on improving word embedding performance in downstream tasks has explored a number of different directions.
Levy and Goldberg \cite{Levy2014c} utilized syntactic dependencies as context,
and found improved functional similarity and decreased topical sensitivity.  
Morphological information about target and context
words has also been incorporated to improve language model robustness in a number of studies, both at the word level
\cite{Luong2013,Cotterell2015,Xu2017a} and implicitly at the character level \cite{Kim2016b}.
Additionally, multilingual data has been extensively investigated for improving language models: Upadhyay et al.\ \cite{
Upadhyay2016} review several recent methods with multilingual corpora, while Faruqui and Dyer \cite{Faruqui2014} and Lu
et al.\ \cite{Lu2015} used canonical correlation analysis to learn cross-lingual embeddings.
Recently, Vuli\'{c} \cite{Vulic2017} combined multilingual
corpora with syntactic dependencies in embedding training, and observed improvements in both monolingual and cross-lingual tasks.

There has also been significant interest in augmenting contextual embedding learning with
information critical to specific downstream tasks.  Faruqui et al.\ \cite{Faruqui2015}, Mrk\u{s}i\'{c} et al.\ \cite{Mrksic2016},
and Kim et al.\ \cite{KimJK2016} enrich pre-trained embeddings with semantic knowledge via lexical constraints.  
Tsvetkov et al.\ \cite{Tsvetkov2016} find benefits from tailoring the learning curriculum or embedding training to specific downstream
tasks, and Rothe et al.\ \cite{Rothe2016} project pre-trained embeddings into task-specific subspaces.  Finally, Yatbaz et
al.\ \cite{Yatbaz2012} use second-order contexts in the form of possible lexical substitutions for word representations;
Melamud et al.\ \cite{Melamud2014,Melamud2015a,Melamud2016} adapt this
approach for embedding learning, by using lexical substitutions to incorporate the joint contexts of two words and extend contextual
information in training.

\subsection{Absolute positioning in the embedding space}

There has also been significant research investment in analyzing and interpreting neural models for NLP, and word embeddings in particular.
Several studies have found correlations between individual dimensions in a sample embedding set and semantic groupings
\cite{Fyshe2015,Faruqui2015b}, as well as with predictive outcomes \cite{Li2016,Li2017a}.  However, in line with the random
initializations used in most embedding training, correlations of specific features vary between studies, and direct interpretability
remains elusive.

Additionally, Li et al.\ \cite{LiY2016} demonstrated sensitivity of neural NLP models to noise in the input space, and present
regularization methods for compositional models to more robustly handle perturbations in input.  Several studies have also shown
wide variability in the reliability of semantic and syntactic information as encoded linearly in the vector space \cite{Chiu2016a,
Gladkova2016,Drozd2016}.  Linzen \cite{Linzen2016} illustrated that many of the successes on similar tasks have relied
more on nearest neighborhood structure than consistent affine transformations.
\section{Second-order embeddings}
\label{sec:methods}

We present a method to generate second-order word embeddings from the nearest neighborhood structure calculated over a
pre-trained embedding vocabulary.
Given a set of pre-trained word embeddings $V$, let $NN^k_v$ denote the $k$ nearest neighbors of each word $v \in V$, as
calculated by cosine similarity.  We then use these $k$ nearest neighbor sets to induce a graph $G_V$ over the vocabulary,
where each word $v$ is a vertex, and the directed edge $(v,w)$ is added for each $w \in NN^k_v$.\footnote{Since we consider only the
$k$ nearest neighbors for each vertex, this may result in graph components that are only connected in one direction.}  An
example induction is shown in Figure~\ref{fig:nn_graph}.  Finally, we use node2vec \cite{Grover2016}, a recent method for
learning unsupervised embeddings of graphs nodes based on weighted random walks, to learn second-order embeddings
for each word in the vocabulary.  This yields a new embedding for each word, based solely on nearest neighbor topological features
from the first-order embedding space.

\begin{figure}
    \tikzset{
        emb img/.style={
            text width=4.5cm,
            minimum height=4.5cm,
            align=center
        },
        vertex/.style={
        },
        edge/.style={
            ->
        }
    }
    \begin{tikzpicture}
        \node [emb img] (emb) {
            \includegraphics[width=\textwidth]{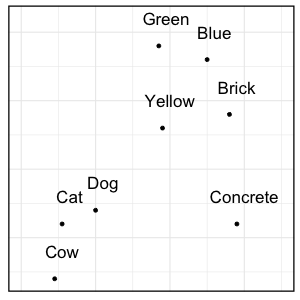}
        };

        \draw [->, gray!80, ultra thick] ($(emb.east) + (0.5,0)$) -- ($(emb.east) + (2.5,0)$);
        \node [right of=emb, node distance=3.85cm, yshift=0.5cm, text width=1.5cm, align=center, font=\it] {3-NN};

        \node [vertex, right of=emb, node distance=9cm] (yellow) {Yellow};
        \node [vertex, above of=yellow, node distance=1.0cm] (blue) {Blue};
        \node [vertex, right of=yellow, node distance=1.5cm, yshift=0.4cm] (green) {Green};
        \node [vertex, below of=yellow, node distance=1.5cm, xshift=-0.5cm] (dog) {Dog};
        \node [vertex, left of=dog, node distance=1.5cm, yshift=-0.25cm] (cat) {Cat};
        \node [vertex, right of=dog, node distance=1.0cm, yshift=-0.35cm] (cow) {Cow};
        \node [vertex, above of=yellow, node distance=1cm, xshift=-2.5cm] (brick) {Brick};
        \node [vertex, below of=brick, node distance=1cm, xshift=-0.5cm] (concrete) {Concrete};

        \draw [edge] (yellow) -- (blue);
        \draw [edge] (yellow) -- (green);
        \draw [edge] (yellow) -- (brick);

        \draw [edge] (green) -- (blue);
        \draw [edge] (green) -- (yellow);
        \draw [edge] (green) -- (brick);

        \draw [edge] (blue) -- (green);
        \draw [edge] (blue) -- (yellow);
        \draw [edge] (blue) -- (brick);

        \draw [edge] (concrete) -- (yellow);
        \draw [edge] (concrete) -- (dog);
        \draw [edge] (concrete) -- (brick);

        \draw [edge] (brick) -- (yellow);
        \draw [edge] (brick) -- (blue);
        \draw [edge] (brick) -- (green);

        \draw [edge] (dog) -- (yellow);
        \draw [edge] (dog) -- (cat);
        \draw [edge] (dog) -- (cow);

        \draw [edge] (cow) -- (cat);
        \draw [edge] (cow) -- (dog);
        \draw [edge] (cow) -- (yellow);

        \draw [edge] (cat) -- (dog);
        \draw [edge] (cat) -- (cow);
        \draw [edge] (cat) -- (yellow);
    \end{tikzpicture}
    \caption{Example induction of a 3-nearest neighbor graph over an embedded vocabulary, using Euclidean distance.  Note that some
    components may not be accessible from other components, e.g., $concrete$ is inaccessible from any other vertex.}
    \label{fig:nn_graph}
\end{figure}
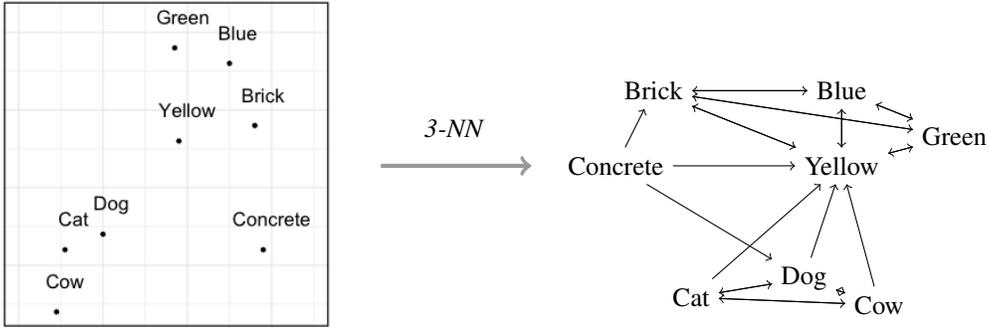
\begin{table}
    \small
    \caption{The 10 nearest neighbors to {\it zucchini}, from three sets of word embeddings trained for 10 iterations via skip-gram
    over a portion of English Gigaword.  Words common to all three samples are marked in bold; italicized words are present
    in only two of the three samples.  Terms are shown in order of increasing distance (column 1 is items 1-5, column 2 is 6-10).}
    \label{tbl:nn_samples}
    \centering
    \vspace{0.3cm}
    \begin{tabular}{cc|cc|cc}
        \multicolumn{2}{c}{Sample 1}&\multicolumn{2}{|c}{Sample 2}&\multicolumn{2}{|c}{Sample 3}\\
        \hline
        {\bf eggplant}&{\bf garlic}&{\bf eggplant}&{\bf roasted}&{\bf eggplant}&{\bf garlic}\\
        {\bf celery}&{\it thyme}&{\bf sauce}&escarole&{\bf sauce}&{\it thyme}\\
        {\bf roasted}&broth&{\bf oregano}&pepperoncini&{\bf celery}&{\bf sorbet}\\
        {\bf oregano}&{\bf sorbet}&{\bf celery}&{\it pancetta}&{\bf oregano}&{\it pancetta}\\
        {\bf sauce}&pesto&{\bf sorbet}&{\bf garlic}&{\bf roasted}&cucumber\\
    \end{tabular}
\end{table}
As most unsupervised embedding methods use a random initialization, the nearest neighborhood structures may vary between
multiple embedding sets; Table~\ref{tbl:nn_samples} shows empirical observations from three embedding samples trained with
skip-gram on English Gigaword 5 \cite{Gigaword5}.  However, our graph induction step can be adjusted to accommodate multiple
samples of nearest neighborhoods for each word, to make it more robust to random initializations.  With sample embedding sets
$\mathcal{V}$, we can calculate the nearest neighborhood $NN^k_{v,i}$ for each word $v$ in each sample $i$.  The weighted edges
in $G$ originating at $v$ are then defined as
\begin{equation*}
    \text{weight}(v,w) = \frac{1}{|\mathcal{V}|}\sum_i f(v,w;i)
\end{equation*}
where $w$ is an element of $\cup_i NN^k_{v,i}$, and $f(v,w;i)$ is an indicator function that returns 1 if $w \in NN^k_{v,i}$ and
0 otherwise.  While both the graph induction and second embedding steps increase the hyperparameters to consider in the model,
the ability to generate a weighted multi-sample nearest neighborhood graph allows for including first-order embeddings trained
with a variety of hyperparameter settings.
\section{Experiments}
\label{sec:experiments}

In order to explore the properties of second-order embeddings, we apply them as input features to proven algorithms for three
tasks: named entity recognition (NER), recognizing textual entailment, and paraphrase recognition.  For all three tasks, we use
existing methods that use word embeddings as input features; for the first two tasks, we use deep and highly non-linear neural
models, while the model for paraphrase recognition is a simple logistic regression.  
Finally, we compare
the nearest neighborhood structure of first-order and second-order embeddings.

To control for
corpus and hyperparameter effects in the different tasks, we use the same sets of embeddings in all applications.  
Our initial word embeddings are trained on Gigaword; following Lample et al.\ \cite{Lample2016}, we remove
the New York Times and LA Times portions of the corpus, and train skip-gram embeddings with word2vec \cite{Mikolov2013a} for 10
iterations, with vector dimensionality of 100, window size of 8, and minimum word frequency of 4.  To induce the nearest neighbor
graph, we choose $k=10$, and then train second-order embeddings with node2vec, again using an embedding size of 100; all other
settings are node2vec defaults.  We also experiment with using three samples in the graph induction
step, as well as varying the neighborhood size $k$ to $k \in \{5,25\}$.

\subsection{Named entity recognition}
We first evaluate our embeddings on the English NER data from the well-studied CoNLL 2003 shared task
\cite{CoNLL03}.  The goal of the task is to take as input unannotated documents and tag within them mentions of persons (PER), locations
(LOC), organizations (ORG), or other entities that do not fit in any of these three categories (MISC).

We adopt the NER system of Lample et al.\ \cite{Lample2016}, which is based on a bidirectional long short-term memory (LSTM)
network with a conditional random field (CRF) over the output layer.\footnote{
    We use their publicly-available implementation: {\tt https://github.com/glample/tagger}
}  They use character embeddings, learned during training, in addition to pre-initialized word embeddings; for our experiments,
we vary the input word embeddings, but do not change character-level behavior.

\begin{table}
    \captionof{table}{Overall results on the CoNLL-03 NER task; highest performing features are marked in bold.  node2vec$_NN$ refers to second-order embeddings with a single sample; $NN^3$ denotes the 3-sample case.  Concatenated refers to concatenating first-order word2vec embeddings with various second-order embeddings.  Pr = Precision, Rec = Recall.}
    \label{tbl:conll_results_overall}
    \centering
    \vspace{0.3cm}
    \begin{tabular}{l|ccc}
        Setting&Pr&Rec&F1\\
        \hline
        word2vec&{\bf 87.65}&{\bf 87.46}&{\bf 87.56}\\
        node2vec$_{NN}$&86.19&87.13&86.66\\
        Concatenated&87.45&{\bf 87.46}&87.46\\
        node2vec$_{NN^3}$&86.94&86.02&86.48\\
        Concatenated ($NN^3$)&87.53&86.75&87.14\\
        Concatenated ($k=5$)&87.30&86.75&87.02\\
        Concatenated ($k=25$)&86.65&86.88&86.76\\
    \end{tabular}
\end{table}
Table~\ref{tbl:conll_results_overall} gives precision, recall, and F-score results for the full test set.  The second-order embeddings
alone perform nearly as well as word2vec embeddings, achieving 1.5\% absolute lower precision, and only 0.5\% absolute lower recall.
Interestingly, using multiple embedding samples to generate the graph increases precision to nearly 87\% at the expense of recall (a
drop of over 1\% absolute compared to the single sample).  Concatenating skip-gram and second-order embeddings gives overall performance
similar to skip-gram alone, with slightly lower precision and identical recall (when using a single sample in graph induction) or
decreased (with multiple samples or smaller neighborhoods).  Increasing the neighborhood size to $k=25$ appears to introduce additional
noise in the neighborhood graph, and decreases performance by a point across the board.

The picture gets more interesting when the results are broken down by named entity type, shown in Table~\ref{tbl:conll_results_typed}.
Most strikingly, nearest neighbor information is critical for recognizing MISC entities, leading to 8.4\% absolute increase in precision
when using multi-sample second order embeddings alone, and similar increases with concatenation.  Recall on MISC falls slightly when
incorporating second-order information, though the noisy $k=25$ graph does increase recall by 0.2\% absolute.  The converse pattern
emerges for PER entities, where precision falls by over a point when including any second-order information, but recall increases
slightly with single-sample nearest neighbor information.  ORG and LOC are less clear, with small variations in precision and recall.

\begin{table}
    \footnotesize
    \caption{Results on CoNLL-03 NER task by named entity type. $n$ is the number of entities of each type in the gold dataset.  Highest performance in each metric on each entity type is marked in bold.}
    \label{tbl:conll_results_typed}
    \centering
    \vspace{0.3cm}
    \begin{tabular}{l|ccc|ccc|ccc|ccc}
        \multirow{2}{*}{Setting}&\multicolumn{3}{c|}{Loc ($n=1668$)}&\multicolumn{3}{c|}{Misc ($n=702$)}&\multicolumn{3}{c|}{Org ($n=1661$)}&\multicolumn{3}{c}{Per ($n=1617$)}\\
        &Pr&Rec&F1&Pr&Rec&F1&Pr&Rec&F1&Pr&Rec&F1\\
        \hline
        word2vec&{\bf 92.3}&91.1&{\bf 91.7}&68.7&79.0&73.5&86.4&{\bf 85.2}&{\bf 85.8}&{\bf 94.0}&89.6&{\bf 91.8}\\
        node2vec$_{NN}$&89.8&91.9&90.8&74.4&78.3&76.3&85.4&82.0&83.7&88.5&91.3&89.9\\
        Concat&91.0&91.9&91.5&72.2&78.0&75.0&{\bf 86.8}&83.2&85.0&91.6&{\bf 91.4}&91.5\\
        node2vec$_{NN^3}$&91.9&90.8&91.4&{\bf 77.1}&77.2&{\bf 77.1}&82.7&82.9&82.8&90.6&88.1&89.3\\
        Concat ($NN^3$)&90.1&91.5&90.8&75.6&76.2&75.9&85.2&84.8&85.0&92.7&88.5&90.5\\
        Concat ({\scriptsize $k=5$})&92.0&90.0&91.0&75.0&77.8&76.3&84.5&84.6&84.5&91.0&89.5&90.3\\
        Concat ({\scriptsize $k=25$})&91.4&{\bf 92.0}&91.7&69.7&{\bf 79.2}&74.2&85.4&82.6&84.0&91.4&89.4&90.4\\
    \end{tabular}
\end{table}

\subsection{Recognizing textual entailment}
For textual entailment, we use the Stanford Natural Language Inference (SNLI) dataset \cite{SNLI}.  The dataset consists of 570,152 sentence
pairs (550,152 for training, 10k for development, and 10k for testing), each of which is annotated with the label {\it entailment} (sentence 1
entails sentence 2), {\it contradiction} (sentence 2 contradicts sentence 1), or {\it neutral} (sentence 1 does not inform sentence 2).  The
three labels are roughly equally distributed.

To evaluate our embeddings on this task, we use the Long Short-Term Memory Network model proposed by Cheng et al.\ \cite{Cheng2016}, which
they evaluated on several machine reading tasks, including SNLI.  We use the publicly-available implementation of their
model,\footnote{
    \tt https://github.com/cheng6076/SNLI-attention
} and compare second-order embeddings with skip-gram embeddings as pre-initialization.  We use their implementation's
default settings of a batch size of 40, sentence embedding dimensionality of 450, dropout of 0.5, and learning rate of 0.001, and vary
the dimensionality of the input word representations to match our embedding settings.  We halt training after 3 iterations.

Table~\ref{tbl:snli_results} presents accuracy of the various models over the development and test sets.  In contrast to the NER
results, here we see an increase in overall performance when augmenting skip-gram embeddings with second-order information.
The second-order embeddings alone give only slightly worse results than the skip-gram embeddings, with around a 1\% drop
in test accuracy.  Interestingly, concatenating skip-gram with a single $k=10$ second-order embedding decreases performance, but using
multiple samples or different settings for $k$ show equivalent or superior performance.

\begin{table}
    \captionof{table}{Accuracy on SNLI development and test sets.}
    \label{tbl:snli_results}
    \centering
    \vspace{0.3cm}
    \begin{tabular}{l|cc}
        Setting&Acc \% (Dev)&Acc \% (Test)\\
        \hline
        word2vec&82.66&82.34\\
        node2vec$_{NN}$&80.85&81.15\\
        Concatenated&82.53&81.87\\
        node2vec${NN^3}$&81.68&81.43\\
        Concatenated ($NN^3$)&82.59&{\bf 82.70}\\
        Concatenated ($k=5$)&{\bf 82.78}&82.64\\
        Concatenated ($k=25$)&82.23&82.34\\
    \end{tabular}
\end{table}
\subsection{Paraphrase recognition}
We also evaluate our embeddings on the task of paraphrase recognition: given two sentences, the task is to decide if sentence 2 is
a paraphrase of sentence 1 or not.  We use the well-studied Microsoft Research Paraphrase Corpus (MSRPC) \cite{MSRPC}, consisting of 5,801 sentence
pairs (4,076 for training, 1,725 for test), each of which is labeled as ``equivalent'' (3,900 pairs, 67\%) or ``not equivalent.''

We follow the methodology of Blacoe and Lapata \cite{Blacoe2012a} for this task.  Specifically, we represent each sentence as the sum\footnote{
    We also experimented with the alternative point-wise multiplication method they discuss, but the low magnitude of our embeddings made it
    impractical for all but the shortest sentences.  For additive composition, our results mirror their findings with neural language models.}
of the embeddings of its in-vocabulary words; the feature vector for a sentence pair is then either the concatenation
or difference of the two sentence embeddings.  For classification, we use logistic regression as implemented in LIBLINEAR \cite{LIBLINEAR}, with
a cost parameter of 0.001.\footnote{
    This cost was empirically determined by experimenting on a validation set created by holding out a randomly-selected 20\% of the training
    data.  We experimented with $c$ from 1 to $10^{-5}$, and found the best performance for first-order, second-order, and concatenated embeddings
    with 0.001.
}

Table~\ref{tbl:msrpc_results} shows precision, recall, and F1 for each embedding set on the MSRPC test set, under the concatenation and subtraction
sentence combination schema.  With concatenated feature vectors, precision varies slightly from the first-order baseline in the various concatenated
settings, but decreases by 1\% absolute when using second-order embeddings alone.  Interestingly, recall behaves inversely: it decreases
significantly in the concatenated settings, but increases when using second-order embeddings alone.  The impact of second-order information in
the subtractive combination scheme is less clear; as with concatenation, precision drops with second-order embeddings alone, but recall is most
strongly affected by using multiple samples in the graph induction step.

\begin{table}
    \footnotesize
    \captionof{table}{Precision, recall, and F-score on the MSRPC paraphrase recognition dataset, using the concatenation
             and subtraction combination methods for sentence embeddings.  Highest performing settings are marked in bold.}
    \label{tbl:msrpc_results}
    \centering
    \vspace{0.3cm}
    \begin{tabular}{l|ccc|ccc}
        \multirow{2}{*}{Setting}&\multicolumn{3}{c}{Concat}&\multicolumn{3}{|c}{Subtract}\\
        &Pr&Rec&F1&Pr&Rec&F1\\
        \hline
        word2vec&69.7&93.2&{\bf 79.8}&{\bf 68.5}&50.6&58.2\\
        node2vec$_{NN}$&68.7&94.0&79.4&66.2&50.3&57.2\\
        Concatenated&69.8&90.7&78.9&68.4&50.7&58.3\\
        node2vec$_{NN^3}$&68.5&{\bf 94.2}&79.3&67.3&{\bf 51.3}&58.2\\
        Concatenated ($NN^3$)&{\bf 69.9}&91.4&79.2&68.1&51.1&{\bf 58.4}\\
        Concatenated ($k=5$)&69.5&91.0&78.8&67.7&50.4&57.8\\
        Concatenated ($k=25$)&69.9&92.2&79.5&68.4&49.7&57.6\\
    \end{tabular}
\end{table}

\subsection{Neighborhood analysis}
\label{subsec:neighborhood_analysis}

Table~\ref{tbl:nn_samples} illustrated the variance in nearest neighborhoods between multiple contextual embedding samples.  Since nearest
neighborhoods remain important in second-order embeddings, we analyzed how these neighborhoods changed from first-order to second-order
representations.  Table~\ref{tbl:nn_samples_n2v} shows the 10 nearest neighbors for {\it zucchini} in second-order embedding spaces
induced from the samples used in Table~\ref{tbl:nn_samples}, along with a space induced from combining all three samples.  While the theme
of cooking remains the same between the two, and both include a number of ingredients appropriate to combine with zucchini, the second-order
samples put more broadly related words near to {\it zucchini}, as opposed to words with highly similar usage patterns.

\begin{table}
    \small
    \caption{The 10 nearest neighbors to {\it zucchini}, in second-order embeddings induced from the three sets of word embeddings from
    Table~\ref{tbl:nn_samples}.  The fourth column corresponds to second-order embeddings using combined nearest neighborhoods from all three
    word embedding samples.  Words common to all four samples are underlined; those present in three samples are marked in bold; and italicized
    words are present in only two of the three samples.  Terms are shown in order of increasing distance.}
    \label{tbl:nn_samples_n2v}
    \centering
    \vspace{0.3cm}
    \begin{tabular}{cc|cc|cc|cc}
        \multicolumn{2}{c}{Sample 1}&\multicolumn{2}{|c}{Sample 2}&\multicolumn{2}{|c}{Sample 3}&\multicolumn{2}{|c}{Multi-sample}\\
        \hline
        \underline{\bf coriander}&oregano&\underline{\bf scallions}&marinate&\underline{\bf scallions}&{\it figs}&{\it cumin}&eggplant\\
        \underline{\bf scallions}&{\it cumin}&\underline{\bf cilantro}&\underline{\bf sprigs}&yolks&jalapenos&\underline{\bf sprigs}&\underline{\bf cinnamon}\\
        \underline{\bf cilantro}&\underline{\bf cinnamon}&applesauce&{\it onion}&\underline{\bf cilantro}&pecans&\underline{\bf cilantro}&grated\\
        {\it figs}&shallots&\underline{\bf coriander}&marinade&{\it chives}&\underline{\bf coriander}&\underline{\bf scallions}&teaspoons\\
        \underline{\bf sprigs}&{\it onion}&\underline{\bf cinnamon}&chickpeas&\underline{\bf cinnamon}&\underline{\bf sprigs}&{\it chives}&\underline{\bf coriander}\\
    \end{tabular}
\end{table}
\begin{table}
    \scriptsize
    \caption{10 nearest neighbors to {\it cibber} and {\it 1976\_ferrari}, in first-order word2vec space and second-order embedding space.  Terms in both first and second-order nearest neighbor lists are bold.}
    \label{tbl:nn_extremal_comparison}
    \centering
    \vspace{0.3cm}
    \begin{tabular}{cc|cc|cc|cc}
        \multicolumn{4}{c}{\it cibber}&\multicolumn{4}{|c}{\it 1976\_ferrari}\\
        \multicolumn{2}{c}{First-order}&\multicolumn{2}{|c}{Second-order}&\multicolumn{2}{|c}{First-order}&\multicolumn{2}{|c}{Second-order}\\
        \hline
        specatators&havoco&impala&disclaimer&{\bf 1975\_ferrari}&{\bf 1973\_lotus}&{\bf 1974\_mclaren}&{\bf 1973\_lotus}\\
        barabati&pramadasa&vanden&repels&{\bf 1977\_ferrari}&{\bf 1972\_lotus}&{\bf 1977\_ferrari}&{\bf 1971\_tyrell}\\
        lutzinger&3ws&bodart&sf/nvw&{\bf 1974\_mclaren}&{\bf 1971\_tyrell}&{\bf 1979\_ferrari}&{\bf 1978\_lotus}\\
        kapaso&overnight&hectolitres&kohlman&{\bf 1979\_ferrari}&1980\_williams&{\bf 1975\_ferrari}&{\bf 1981\_williams}\\
        snapping&a\&t\_announced&schedule&nder&{\bf 1978\_lotus}&{\bf 1981\_williams}&{\bf 1972\_lotus}&1970\_lotus\\
    \end{tabular}
\end{table}
\begin{figure}
    \begin{minipage}{.5\textwidth}
        \centering
\begin{tikzpicture}[x=1pt,y=1pt]
\definecolor{fillColor}{RGB}{255,255,255}
\path[use as bounding box,fill=fillColor,fill opacity=0.00] (0,0) rectangle (180.67,144.54);
\begin{scope}
\path[clip] ( 36.71, 16.51) rectangle (175.17,139.04);
\definecolor{drawColor}{gray}{0.92}

\path[draw=drawColor,line width= 0.3pt,line join=round] ( 36.71, 32.75) --
	(175.17, 32.75);

\path[draw=drawColor,line width= 0.3pt,line join=round] ( 36.71, 67.03) --
	(175.17, 67.03);

\path[draw=drawColor,line width= 0.3pt,line join=round] ( 36.71,101.30) --
	(175.17,101.30);

\path[draw=drawColor,line width= 0.3pt,line join=round] ( 36.71,135.58) --
	(175.17,135.58);

\path[draw=drawColor,line width= 0.6pt,line join=round] ( 36.71, 49.89) --
	(175.17, 49.89);

\path[draw=drawColor,line width= 0.6pt,line join=round] ( 36.71, 84.16) --
	(175.17, 84.16);

\path[draw=drawColor,line width= 0.6pt,line join=round] ( 36.71,118.44) --
	(175.17,118.44);

\path[draw=drawColor,line width= 0.6pt,line join=round] ( 74.47, 16.51) --
	( 74.47,139.04);

\path[draw=drawColor,line width= 0.6pt,line join=round] (137.41, 16.51) --
	(137.41,139.04);
\definecolor{drawColor}{gray}{0.20}
\definecolor{fillColor}{gray}{0.20}

\path[draw=drawColor,line width= 0.4pt,line join=round,line cap=round,fill=fillColor] ( 74.47, 35.26) circle (  1.96);

\path[draw=drawColor,line width= 0.4pt,line join=round,line cap=round,fill=fillColor] ( 74.47, 34.26) circle (  1.96);

\path[draw=drawColor,line width= 0.4pt,line join=round,line cap=round,fill=fillColor] ( 74.47, 40.14) circle (  1.96);

\path[draw=drawColor,line width= 0.4pt,line join=round,line cap=round,fill=fillColor] ( 74.47, 35.62) circle (  1.96);

\path[draw=drawColor,line width= 0.4pt,line join=round,line cap=round,fill=fillColor] ( 74.47, 34.71) circle (  1.96);

\path[draw=drawColor,line width= 0.6pt,line join=round] ( 74.47, 28.72) -- ( 74.47, 33.36);

\path[draw=drawColor,line width= 0.6pt,line join=round] ( 74.47, 25.20) -- ( 74.47, 22.08);
\definecolor{fillColor}{RGB}{248,118,109}

\path[draw=drawColor,line width= 0.6pt,line join=round,line cap=round,fill=fillColor] ( 50.87, 28.72) --
	( 50.87, 25.20) --
	( 98.08, 25.20) --
	( 98.08, 28.72) --
	( 50.87, 28.72) --
	cycle;

\path[draw=drawColor,line width= 1.1pt,line join=round] ( 50.87, 27.08) -- ( 98.08, 27.08);
\definecolor{fillColor}{gray}{0.20}

\path[draw=drawColor,line width= 0.4pt,line join=round,line cap=round,fill=fillColor] (137.41,128.57) circle (  1.96);

\path[draw=drawColor,line width= 0.4pt,line join=round,line cap=round,fill=fillColor] (137.41,133.47) circle (  1.96);

\path[draw=drawColor,line width= 0.4pt,line join=round,line cap=round,fill=fillColor] (137.41,124.33) circle (  1.96);

\path[draw=drawColor,line width= 0.4pt,line join=round,line cap=round,fill=fillColor] (137.41,125.92) circle (  1.96);

\path[draw=drawColor,line width= 0.4pt,line join=round,line cap=round,fill=fillColor] (137.41,125.66) circle (  1.96);

\path[draw=drawColor,line width= 0.4pt,line join=round,line cap=round,fill=fillColor] (137.41,124.39) circle (  1.96);

\path[draw=drawColor,line width= 0.4pt,line join=round,line cap=round,fill=fillColor] (137.41,125.17) circle (  1.96);

\path[draw=drawColor,line width= 0.6pt,line join=round] (137.41,117.95) -- (137.41,123.86);

\path[draw=drawColor,line width= 0.6pt,line join=round] (137.41,113.96) -- (137.41,110.36);
\definecolor{fillColor}{RGB}{0,191,196}

\path[draw=drawColor,line width= 0.6pt,line join=round,line cap=round,fill=fillColor] (113.81,117.95) --
	(113.81,113.96) --
	(161.01,113.96) --
	(161.01,117.95) --
	(113.81,117.95) --
	cycle;

\path[draw=drawColor,line width= 1.1pt,line join=round] (113.81,115.57) -- (161.01,115.57);
\definecolor{drawColor}{RGB}{0,0,0}

\path[draw=drawColor,line width= 1.4pt,line join=round,line cap=round] ( 36.71, 16.51) rectangle (175.18,139.04);
\end{scope}
\begin{scope}
\path[clip] (  0.00,  0.00) rectangle (180.67,144.54);
\definecolor{drawColor}{gray}{0.30}

\node[text=drawColor,anchor=base east,inner sep=0pt, outer sep=0pt, scale=  0.88] at ( 31.76, 46.86) {0.4};

\node[text=drawColor,anchor=base east,inner sep=0pt, outer sep=0pt, scale=  0.88] at ( 31.76, 81.13) {0.6};

\node[text=drawColor,anchor=base east,inner sep=0pt, outer sep=0pt, scale=  0.88] at ( 31.76,115.41) {0.8};
\end{scope}
\begin{scope}
\path[clip] (  0.00,  0.00) rectangle (180.67,144.54);
\definecolor{drawColor}{gray}{0.30}

\node[text=drawColor,anchor=base,inner sep=0pt, outer sep=0pt, scale=  0.88] at ( 74.47,  5.50) {First-order};

\node[text=drawColor,anchor=base,inner sep=0pt, outer sep=0pt, scale=  0.88] at (137.41,  5.50) {Second-order};
\end{scope}
\begin{scope}
\path[clip] (  0.00,  0.00) rectangle (180.67,144.54);
\definecolor{drawColor}{RGB}{0,0,0}

\node[text=drawColor,rotate= 90.00,anchor=base,inner sep=0pt, outer sep=0pt, scale=  1.10] at ( 13.08, 77.78) {Cosine similarity};
\end{scope}
\end{tikzpicture}
    \end{minipage}%
    \begin{minipage}{.5\textwidth}
        \centering
\begin{tikzpicture}[x=1pt,y=1pt]
\definecolor{fillColor}{RGB}{255,255,255}
\path[use as bounding box,fill=fillColor,fill opacity=0.00] (0,0) rectangle (180.67,144.54);
\begin{scope}
\path[clip] ( 45.51, 16.51) rectangle (175.17,139.04);
\definecolor{drawColor}{gray}{0.92}

\path[draw=drawColor,line width= 0.3pt,line join=round] ( 45.51, 23.63) --
	(175.17, 23.63);

\path[draw=drawColor,line width= 0.3pt,line join=round] ( 45.51, 48.35) --
	(175.17, 48.35);

\path[draw=drawColor,line width= 0.3pt,line join=round] ( 45.51, 73.07) --
	(175.17, 73.07);

\path[draw=drawColor,line width= 0.3pt,line join=round] ( 45.51, 97.79) --
	(175.17, 97.79);

\path[draw=drawColor,line width= 0.3pt,line join=round] ( 45.51,122.51) --
	(175.17,122.51);

\path[draw=drawColor,line width= 0.6pt,line join=round] ( 45.51, 35.99) --
	(175.17, 35.99);

\path[draw=drawColor,line width= 0.6pt,line join=round] ( 45.51, 60.71) --
	(175.17, 60.71);

\path[draw=drawColor,line width= 0.6pt,line join=round] ( 45.51, 85.43) --
	(175.17, 85.43);

\path[draw=drawColor,line width= 0.6pt,line join=round] ( 45.51,110.15) --
	(175.17,110.15);

\path[draw=drawColor,line width= 0.6pt,line join=round] ( 45.51,134.87) --
	(175.17,134.87);

\path[draw=drawColor,line width= 0.6pt,line join=round] ( 80.87, 16.51) --
	( 80.87,139.04);

\path[draw=drawColor,line width= 0.6pt,line join=round] (139.81, 16.51) --
	(139.81,139.04);
\definecolor{drawColor}{gray}{0.20}
\definecolor{fillColor}{gray}{0.20}

\path[draw=drawColor,line width= 0.4pt,line join=round,line cap=round,fill=fillColor] ( 80.87, 41.63) circle (  1.96);

\path[draw=drawColor,line width= 0.4pt,line join=round,line cap=round,fill=fillColor] ( 80.87, 40.91) circle (  1.96);

\path[draw=drawColor,line width= 0.4pt,line join=round,line cap=round,fill=fillColor] ( 80.87, 43.93) circle (  1.96);

\path[draw=drawColor,line width= 0.4pt,line join=round,line cap=round,fill=fillColor] ( 80.87, 31.99) circle (  1.96);

\path[draw=drawColor,line width= 0.4pt,line join=round,line cap=round,fill=fillColor] ( 80.87, 22.08) circle (  1.96);

\path[draw=drawColor,line width= 0.4pt,line join=round,line cap=round,fill=fillColor] ( 80.87, 37.88) circle (  1.96);

\path[draw=drawColor,line width= 0.6pt,line join=round] ( 80.87,101.80) -- ( 80.87,112.14);

\path[draw=drawColor,line width= 0.6pt,line join=round] ( 80.87, 80.13) -- ( 80.87, 61.65);
\definecolor{fillColor}{RGB}{248,118,109}

\path[draw=drawColor,line width= 0.6pt,line join=round,line cap=round,fill=fillColor] ( 58.77,101.80) --
	( 58.77, 80.13) --
	(102.97, 80.13) --
	(102.97,101.80) --
	( 58.77,101.80) --
	cycle;

\path[draw=drawColor,line width= 1.1pt,line join=round] ( 58.77, 92.55) -- (102.97, 92.55);
\definecolor{fillColor}{gray}{0.20}

\path[draw=drawColor,line width= 0.4pt,line join=round,line cap=round,fill=fillColor] (139.81,131.41) circle (  1.96);

\path[draw=drawColor,line width= 0.4pt,line join=round,line cap=round,fill=fillColor] (139.81,131.61) circle (  1.96);

\path[draw=drawColor,line width= 0.4pt,line join=round,line cap=round,fill=fillColor] (139.81,131.59) circle (  1.96);

\path[draw=drawColor,line width= 0.4pt,line join=round,line cap=round,fill=fillColor] (139.81,131.49) circle (  1.96);

\path[draw=drawColor,line width= 0.6pt,line join=round] (139.81,132.83) -- (139.81,133.47);

\path[draw=drawColor,line width= 0.6pt,line join=round] (139.81,132.36) -- (139.81,131.70);
\definecolor{fillColor}{RGB}{0,191,196}

\path[draw=drawColor,line width= 0.6pt,line join=round,line cap=round,fill=fillColor] (117.71,132.83) --
	(117.71,132.36) --
	(161.91,132.36) --
	(161.91,132.83) --
	(117.71,132.83) --
	cycle;

\path[draw=drawColor,line width= 1.1pt,line join=round] (117.71,132.58) -- (161.91,132.58);
\definecolor{drawColor}{RGB}{0,0,0}

\path[draw=drawColor,line width= 1.4pt,line join=round,line cap=round] ( 45.51, 16.51) rectangle (175.17,139.04);
\end{scope}
\begin{scope}
\path[clip] (  0.00,  0.00) rectangle (180.67,144.54);
\definecolor{drawColor}{gray}{0.30}

\node[text=drawColor,anchor=base east,inner sep=0pt, outer sep=0pt, scale=  0.88] at ( 40.56, 32.96) {0.980};

\node[text=drawColor,anchor=base east,inner sep=0pt, outer sep=0pt, scale=  0.88] at ( 40.56, 57.68) {0.985};

\node[text=drawColor,anchor=base east,inner sep=0pt, outer sep=0pt, scale=  0.88] at ( 40.56, 82.40) {0.990};

\node[text=drawColor,anchor=base east,inner sep=0pt, outer sep=0pt, scale=  0.88] at ( 40.56,107.12) {0.995};

\node[text=drawColor,anchor=base east,inner sep=0pt, outer sep=0pt, scale=  0.88] at ( 40.56,131.84) {1.000};
\end{scope}
\begin{scope}
\path[clip] (  0.00,  0.00) rectangle (180.67,144.54);
\definecolor{drawColor}{gray}{0.30}

\node[text=drawColor,anchor=base,inner sep=0pt, outer sep=0pt, scale=  0.88] at ( 80.87,  5.50) {First-order};

\node[text=drawColor,anchor=base,inner sep=0pt, outer sep=0pt, scale=  0.88] at (139.81,  5.50) {Second-order};
\end{scope}
\begin{scope}
\path[clip] (  0.00,  0.00) rectangle (180.67,144.54);
\definecolor{drawColor}{RGB}{0,0,0}

\node[text=drawColor,rotate= 90.00,anchor=base,inner sep=0pt, outer sep=0pt, scale=  1.10] at ( 13.08, 77.78) {Cosine similarity};
\end{scope}
\end{tikzpicture}
    \end{minipage}
    \caption{Distribution of pairwise similarities between 10 selected words and their 10 nearest neighbors in a first-order
    embedding space and in a corresponding second-order embedding.  Words with lowest mean similarity to their neighbors
    are shown at left, and highest mean similarity at right.}
    \label{fig:similarity_distribs}
\end{figure}
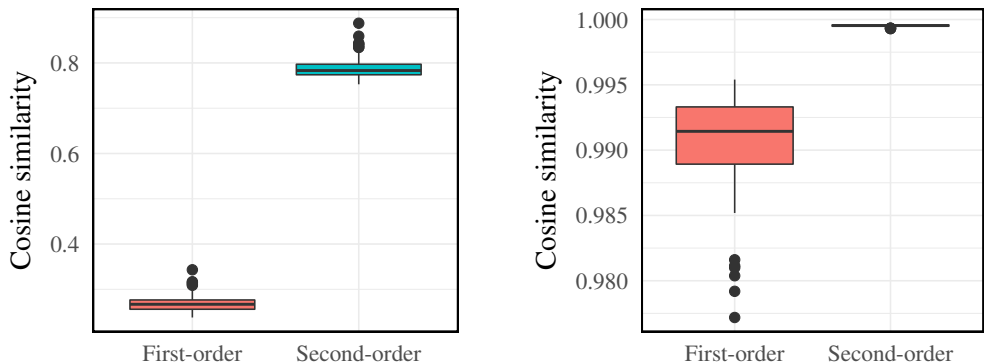

{\it Zucchini} is a reasonably frequent word, occurring 245 times in our Gigaword subset.  However, the nearest neighborhood
graph induction step can be considered as a normalization of distances in the embedding space, in that outliers that had universally
low similarity in the first-order space and words with extremely high similarity to their neighbors both end up connected to $k$
neighbor words in the nearest neighbor graph, at an edge distance of 1.  This raises the question:
what happens to these outliers and dense points in the second-order space?  To answer this question, for each word in the original
embedding space, we calculated the average of its similarity to its 10 nearest neighbors.  Table~\ref{tbl:nn_extremal_comparison}
shows the nearest neighbors in the first and second-order spaces of two words: {\it cibber} (frequency 4; the name of a vocal soloist
mentioned in two articles on a musical performance), which had one of the lowest such average similarities, and
{\it 1976\_ferrari}\footnote{
    Some multi-word expressions in the Gigaword corpus were conjoined with underscores, and these were kept in the plaintext version
    we used for training our embeddings.
} (frequency 10; discussed in the context of Formula 1 racing), which had the highest average similarity.  Qualitatively, unrelated nearest
neighbors in the first-order space lead to unrelated nearest neighbors in the second-order space, while highly related neighbors stay
nearby.

In a quantitative analysis, however, the distance normalization has a much clearer effect.  We took the 10 words in each of the
first-order and second-order embedding spaces with maximal or minimal average similarity to its nearest neighbors, and compiled
the similarities of each word to its 10 nearest neighbors (yielding 100 maximal similarity sample points, and 100 minimal samples).
Figure~\ref{fig:similarity_distribs} shows the distribution of these minimal and maximal similarities in each space.  We see that
the second-order space is strikingly denser than the first-order space: the lowest pairwise similarity of a word to one of its
nearest neighbors in the second-order space is 0.75, in contrast to 0.24\ in the first order space.  Moreover, already dense areas
of the first-order space become appreciably denser, with all 100 of the top pairwise similarities clustering around 0.999.
\section{Discussion}
\label{sec:discussion}

Second-order embeddings retain a surprisingly high degree of the discriminative signal encoded by first-order embeddings,
as measured by their performance on our semantic tasks.  The consistency of second-order embedding performance relative to
first-order embeddings, typically only differing by 1 to 2 points absolute in both the deep and linear models, suggests that
the nearest neighborhood topology of an embedding space contains the lion's share of the important information for these
tasks, independent of the values of individual features.  This holds true in spite of the greater density of second-order
embeddings, and the different semantic correlations we observe between nearby points.

However, a couple of specific performance differences stand out when comparing non-concatenated first-order and second-order
embeddings.  On the NER task,
the distinct increase in precision for MISC entities (which are highly heterogeneous in their textual realizations), and a
large drop in precision for the much more consistent PER category, suggest that the second-order embeddings handle variance in
textual patterns more effectively, but with a corresponding loss of specificity.  This is reflected as well in the paraphrase
recognition results, where second-order representations trade lower precision for higher recall.

Furthermore, concatenating the different embeddings poses an interesting challenge for use in downstream tasks.
The large drop in performance when using concatenated vectors in the linear model for paraphrase recognition indicates that not
only do the second-order embeddings contain different information from the first-order vectors, but that aligning the two
sets of signals with a linear transformation is a challenge.  The highly non-linear models for NER and textual entailment, on
the other hand, can adjust to the combined space, and in the latter case even benefit from the dual signals.

\section{Conclusion}
\label{sec:conclusion}

We introduced second-order word embeddings, derived from the nearest neighborhood topology of context-based word representations.
We analyzed the effects of using these embeddings in existing models for named entity recognition, recognizing textual entailment,
and paraphrase recognition, both as the second-order information alone and concatenated with the first-order contextual embeddings.
Our analysis demonstrated that second-order embeddings yield similar performance to their first-order counterparts, often trading some
specificity (reflected in decreased precision values) for improved handling of heterogeneous data (reflected in increased recall).
Furthermore, we illustrated that the second-order embedding space is much denser than its first-order version, and we found that high
second-order similarity is more indicative of broad relatedness than contextual similarity.

Our findings suggest that second-order embeddings are an intriguing area for further research.  In particular, the higher recall we
observe on all three tasks indicates that second-order embeddings contain valuable information for reliably dealing with heterogeneous
data.  It is clear that non-linear transformations help in combining this information with the direct contextual signals from first-order
embeddings, but how best to find that combination remains an open question.
Additionally, the ability to derive a single second-order representation from multiple samples, and the often superior performance
achieved by doing so,
suggests that this method could be used to reduce some of the variance we observe between different contextual embedding samples
trained on the same data.

\subsubsection*{Acknowledgments}

The authors would like to thank Adam Stiff and Chunxiao Zhou for helpful discussions, as
well as the Ohio Supercomputer Center \cite{OSC} for use of experimental resources.

Denis is a pre-doctoral fellow at the National Institutes of Health, Clinical Center.

\bibliographystyle{unsrt}
\bibliography{references.clean}

\end{document}